\documentclass[12pt,a4paper,titlepage]{article}
\usepackage{amsmath}
\usepackage{amsfonts}

\usepackage{float}
\usepackage{setspace}
\usepackage{subfig}
\usepackage{cite}
\usepackage{amssymb}
\newcommand{\HRule}{\rule{\linewidth}{0.5mm}}
\usepackage{hyperref}
\usepackage{bm}
\usepackage[]{todonotes}
\usepackage{amsmath}
\usepackage[all]{hypcap}
\usepackage[hyperpageref]{backref}
\hypersetup{colorlinks=true,linkcolor=red,citecolor=blue,pdfborder=0 0 0}

\begin{document}

\begin{titlepage}

\begin{center}

\HRule \\[0.3cm]

{\begin{spacing}{1} \huge \bfseries \textcolor{violet} {Analysis and Recognition in Images and Video \normalsize
\textcolor{black} {
\\[0.5cm]
{\huge{Face Recognition using Curvelet Transform}}\\
\vspace{0.3in}
{\huge{Project Report}}\\
}} \end{spacing}}

\HRule \\[1.5cm]

\large
\begin{flushleft}
\emph{Author:}\\
\textcolor{blue} {Rami \textsc{Cohen}} (rc@tx.technion.ac.il)\\[0.3cm]
\normalsize {\textit{This report is accompanied by a MATLAB package that can be requested by mail.}}
\end{flushleft}

\vspace{2in}
{\textcolor{black}{\textbf{Technion - Israel Institute of Technology}}}
\\
{\textcolor{black}{\textbf{Winter 2010/11\\ February 2011}}}

\end{center}

\end{titlepage}

\begin{abstract}

Face recognition has been studied extensively for more than 20 years now. Since the beginning of 90s the subject has became a major issue. This technology is used in many important real-world applications, such as video surveillance, smart cards, database security, internet and intranet access.

This report reviews recent two algorithms for face recognition which take advantage of a relatively new multiscale geometric analysis tool - Curvelet transform, for facial processing and feature extraction. This transform proves to be efficient especially due to its good ability to detect curves and lines, which characterize the human's face.

An algorithm which is based on the two algorithms mentioned above is proposed, and its performance is evaluated on three data bases of faces: AT\&T (ORL), Essex Grimace and Georgia-Tech. k-nearest neighbour (k-NN) and Support vector machine (SVM) classifiers are used, along with Principal Component Analysis (PCA) for dimensionality reduction.

This algorithm shows good results, and it even outperforms other algorithms in some cases.

\end{abstract}
\tableofcontents
\newpage

\section{Introduction}

Over the last ten years or so, face recognition has become a popular area of research in computer vision and one of the most successful applications of image analysis and understanding. Because of the nature of the problem, not only computer science researchers are interested in it, but neuroscientists and psychologists also. It is the general opinion that advances in computer vision research will provide useful insights to neuroscientists and psychologists into how human brain works, and vice versa.

\begin{figure}[h!]
\begin{center}
\includegraphics[scale=0.8]{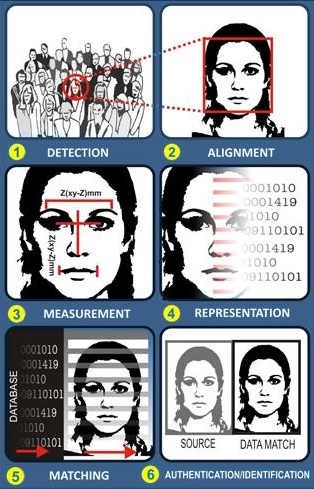}
\end{center}
\caption{General face recognition system}
\label{block_areas}
\end{figure}

General face recognition algorithms include three key steps: (1) face detection and normalization; (2) feature extraction; (3) identification or verification. General recognition process is depicted in Figure \ref{block_areas}. In this project we will focus on the last two steps. 

Facial feature extraction is crucial to face recognition and facial expression recognition. It is clear that an appropriate choice of the representative feature has a crucial effect on the the performance of recognition algorithm.  

Studies in human visual system and image statistics show that an ideal image representation or a feature extraction method should satisfy the following five conditions \cite{Maj07}: multiresolution, localization, critical sampling, directionality and anisotropy.

\begin{figure}[h!]
\begin{center}
\includegraphics[scale=0.5]{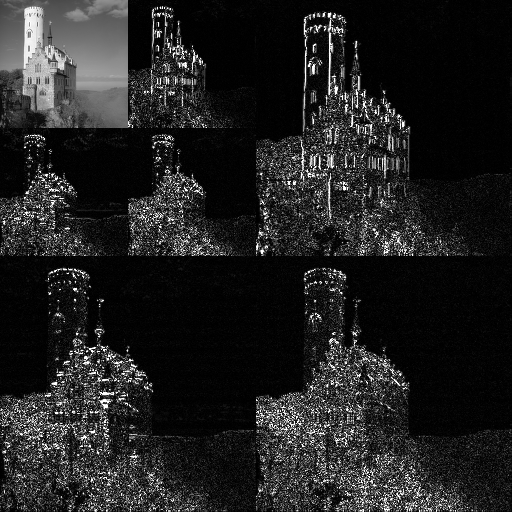}
\end{center}
\caption{Example of the 2D discrete wavelet transform, from \cite{wiki:wt}}
\label{wavelet_transform}
\end{figure}

In recent years, many outstanding algorithms have been proposed for feature extraction. Wavelet analysis (using the well-known wavelet transform) is a significant feature extraction tool because of its ability of localization in both time domain and frequency domain, which can help us in focusing on specific parts of a given image. An example is depicted in Figure \ref{wavelet_transform}.

Wavelet transform is also used in image compression algorithms, such as JPEG-2000. However, wavelet transform can only capture point singularities in an image, rather than curves and lines which appear in the human's face. Moreover, this transform does not satisfy the last two conditions above.

A more suitable transform for the task of face recognition is the Curvelet transform. It is a relatively new mutliscale analysis tool, which was proposed in 1999 \cite{curvelets} and revised in 2006 \cite{FDCT}. Curvelets provide optimally sparse representations of objects which display smoothness except for discontinuity along a general curve with bounded curvature. 

In this project, two algorithms \cite{orig_paper, pca_paper} which use this transform are described and discussed. Both of them use the transform coefficients which are extracted using the curvelet transform, but the second one also employs Principal Component Analysis (PCA) in order to reduce dimensionality. This reduction plays an important role in the process, since a typical $100 \times 100$ pixels image can have thousands of coefficients.

An algorithm which integrates both algorithms above with some improvements is implemented, and its performance are evaluated. It is shown that it produces similar results and in some cases even better results than the algorithms mentioned above.

This report is organized as follows. First, in section \ref{CT} we review the Curvelet transform. Later, we describe and discuss the algorithms in \ref{known_algorithms}. The algorithm which is implemented is described in sections \ref{approach} and \ref{results}, with appropriate performance evaluation, and it is also compared to both algorithm above. Information about the attached Matlab code is given in \ref{matlab_code}. Finally, conclusions are given in section \ref{conclusions}.

Descriptions of k-NN and SVM are presented in Appendices \ref{k-NN_section} and \ref{SVM_section}.

\clearpage
\section{Curvelet transform}
\label{CT}

\begin{figure}[h!]
\begin{center}
\includegraphics[scale=0.4]{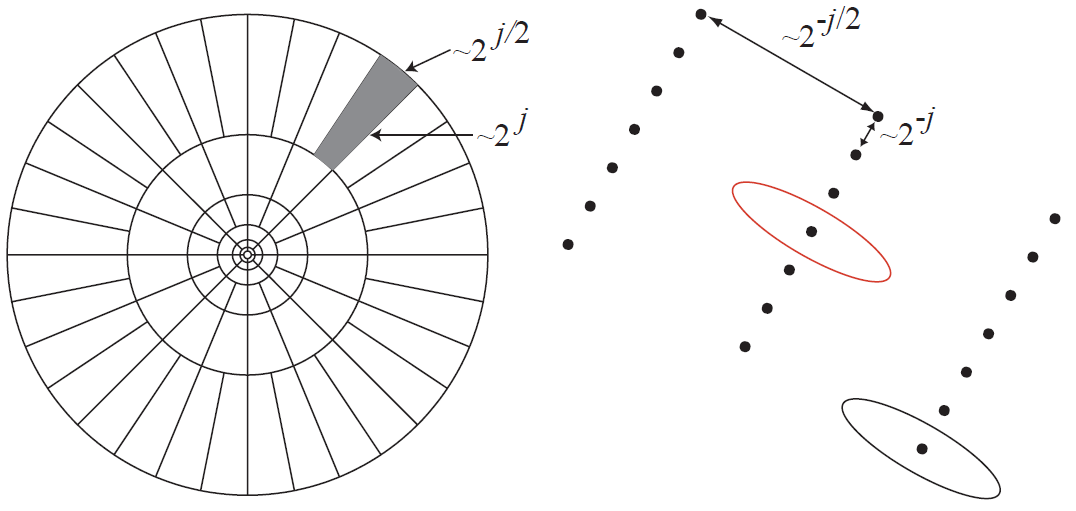}
\end{center}
\caption{Curvelet tiling of space and frequency. The figure on the left represents the induced tiling of the frequency plane The figure on the right schematically represents the spatial Cartesian grid associated with a given scale and orientation}
\label{curvelet_grid}
\end{figure}

Curvelet transform is a kind of multi-resolution analysis tool. Its main advantage is the ability to use relatively small number of coefficients to reconstruct edge details at an image. Each matrix of coefficients is characterized by both an angle and scale. That is, \[C_{m,n}^{j,l} = \left\langle {f,\phi _{m,n}^{j.l}} \right\rangle \]

where $\phi$ is the basis functions, $j$ and $l$ are scale and angle, and $(m,n)$ is an index, which is limited according to $j$ and $l$. Illustration of scale and angle are depicted in figure \ref{curvelet_grid}.

Full description of the curvelet transform and its digital implementations is given in \cite{FDCT}. We will review here one of the implementations - Curvelets via Wrapping, which is faster than the Curvelets via USFFT (Unequally Spaced Fast Fourier Transform) implementation. The transform can be implemented as follows:

\begin{enumerate}

\item{{\bf 2D-FFT}} 2D FFT (Fast Fourier Transform) is applied to obtain Fourier samples: \[\hat f\left[ {{n_1},{n_2}} \right], - n/2 \le {n_1},{n_2} < n/2\]

\item{{\bf Interpolation}} For each scale-angle pair $\left( {j,l} \right)$ interpolate (or resample) $\hat f\left[ {{n_1},{n_2}} \right]$ to obtain $\hat f\left[ {{n_1},{n_2} - {n_1}\tan {\theta _l}} \right]$ (as depicted in figure \ref{curvelet_grid}).

\item{{\bf Localization}} Multiply the interpolated function ${\hat f}$ with a window function ${{\tilde U}_j}\left[ {{n_1},{n_2}} \right]$ effectively localizing ${\hat f}$ near the parallelogram with orientation $\theta_l$, to obtain
\[{{\tilde f}_{j,l}}\left[ {{n_1},{n_2}} \right] = \hat f\left[ {{n_1},{n_2} - {n_1}\tan {\theta _l}} \right]{{\tilde U}_j}\left[ {{n_1},{n_2}} \right]\]

\item{{\bf Inverse 2D-FFT}} Apply the inverse 2D FFT to each ${{\tilde f}_{j,l}}$ to obtain the associated \emph{Curvelets} - $C_{m,n}^{j,l}$. It should be noted that the values can be complex. Usually, we will use their norm.

\end{enumerate}
 
\begin{figure}[h!]
\begin{center}
\includegraphics[scale=0.4]{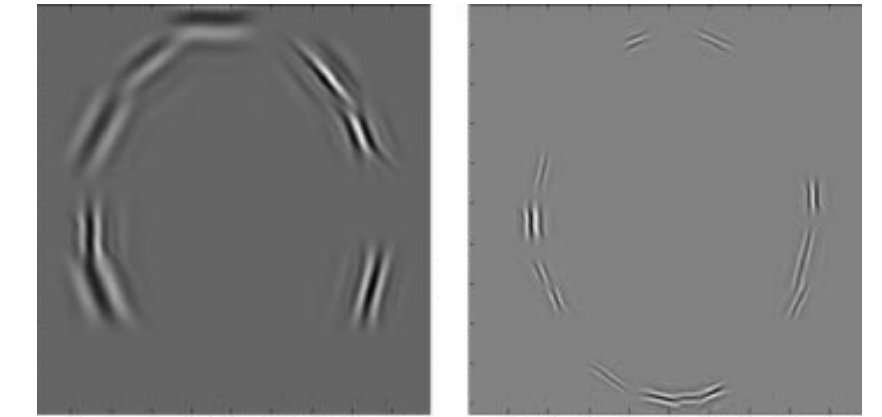}
\end{center}
\caption{A few curvelets (their norms are presented)}
\label{curvelets_example}
\end{figure}

Some curvelets of a circular object can be seen in Figure \ref{curvelets_example}.

Efficient numerical algorithms exist for computing the curvlet transform of discrete data. The computational cost of a curvlet transform is approximately 10-20 times that of an FFT, and has the same dependence of $O(n^2{log(n)})$ for an image of size $n \times n$.

\clearpage
\section{Known algorithms}
\label{known_algorithms}
\subsection{Face Recognition by Curvelet Based Feature Extraction}
\label{orig_algorithm}
\begin{figure}[h!]
\begin{center}
\includegraphics[scale=0.4]{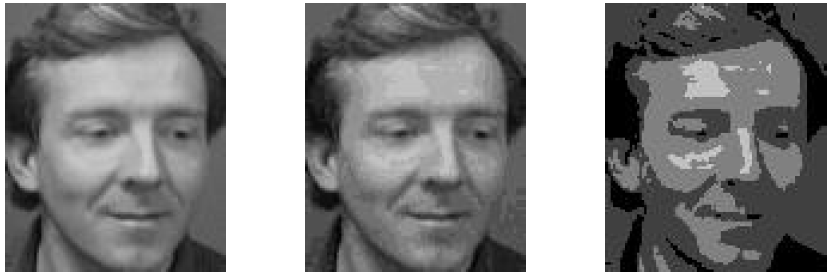}
\end{center}
\caption{Original image (left) and its quantized versions}
\label{image_quanztization}
\end{figure}

This algorithm \cite{orig_paper} uses three different versions of the same image - 8 bit (original), 4 bit and 2 bit. The last two version are obtained by quantization of the original image. This is illustrated in Figure \ref{image_quanztization}. The idea is to find the prominent edges, which will stay even in the quantized image. It uses 3 classifiers whose inputs (features) are the curvelet coefficients of the three gray scale representations of the same image which are mentioned above.

Since there are curvelets for each scale and angle, there is some freedom of the choice of the coefficients (unless all the curvelets are taken into account, which is not the case). However, it is not clear from the description of this algorithm how they should be chosen.

Each testing sample undergoes the same procedure (including quantization), and according to majority vote of the 3 classifiers, its class (person) is decided. If there is no "winner", the image is declared as "rejected". The classifiers which are used are SVMs (Appendix \ref{SVM_section}).

This method has two main drawbacks. First, it is computationally expensive, especially when dealing with large databases, since the transform is done 3 times (on 3 different versions of the same image). Moreover, the assumption that the prominent edges stay even after quantization can be sometimes misleading; actually, some additional contours can be created in the quantization process.

\subsection{Face recognition using curvelet based {PCA} }
\label{pca_algorithm}

This algorithm \cite{pca_paper} works in a similar way to the previous one (\ref{orig_algorithm}), but it does not use the quantized version of the image, and it works directly on the unquantized image. This algorithm introduces significant complexity reduction by the use of PCA. It was shown in this paper that even 50 principal components provide good performance. Moreover, this algorithm provides better results than algorithms which use wavelet transform.

In this algorithm, the classification process is done using the k-NN algorithm, with $k=1$. The coefficients are chosen from one scale (in this paper the scale was chosen to be 3), where all the coefficients which associated with this scale are used (they are inserted to a one long row vector). 

\clearpage
\section{The approach in this project}
\label{approach}

\begin{figure}[h!]
\begin{center}
\includegraphics[scale=0.8]{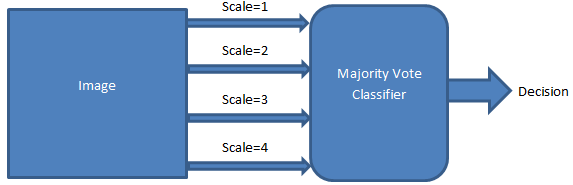}
\end{center}
\caption{Algorithm - General scheme}
\label{algorithm_scheme}
\end{figure}

The approach taken in this project tries to integrate both former algorithms. First, it is clear that the complexity should be tolerable, so transforming different versions of the same image should not be used. Moreover, instead of using the curvelets in only one scale, we can use different classifier for each scale (it should be clear that this still requires only one transform per image), and decide by majority vote on the class of each image. This procedure is depicted in Figure \ref{algorithm_scheme}. Similar idea appears in \cite{5583642}.

The classifiers which are used are k-NN and SVM (Support Vector Machines). Description of these classification algorithms is given in the appendices \ref{k-NN_section}, \ref{SVM_section}. We should also evaluate performance when only small number of pictures is available for training (this is the situation in most scenarios). 

It has been shown in \cite{Pinales} that the recognition accuracy of face images does not degrade significantly if the size of the image is reduced (especially for high resolution pictures). Hence, pictures from AT\&T database (which are $640 \times 480$) were reduced by six times before additional processing. After this operation, their size is similar to images from the other two databases. Moreover, color pictures will be converted to gray scale images.

\begin{figure}[h!]
\begin{center}
\includegraphics[scale=0.8]{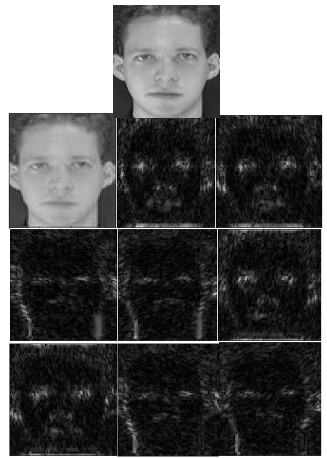}
\end{center}
\caption{The first image is the original image. The left most image in the second row is the approximated image. The rest are curvelets in 8 different angles.}
\label{curvelets_faces}
\end{figure}

The transform involves 8 and 16 angles for scales 2 and 3, respectively, where for scale 1 and 4 no orientation is taken into account (i.e., angle=0). The feature vector per each scale is created by concatenating of all the values in the same scale. Since these values can be complex, we will use only their 2-norm. A face and some of its curvelets are shown in Figure \ref{curvelets_faces}.

It should be noted that SVM classification requires a special treatment, since the basic type of SVM can distinguish between only 2 classes. Hence, we will use One-Against-All (OAA) SVM, i.e., each picture will be classified as belongs to a specific class or not (with no additional information about other classes). More information is given in Appendix \ref{SVM_section}.

Despite the simplicity of k-NN, it provided better results than SVM OAA, so the results which are brought in the next section were obtained by k-NN. 

\clearpage
\section{Results}
\label{results}

In the following parts the results of the approach taken in this project are introduced. They are given for different number of samples in each training set. These results are quite similar for PCA components in the range 50-100 so 100 components are used.

The use of PCA reduces the complexity of the classification part by a few magnitudes. Even small number of coefficients (say, 100 instead of thousands), proves to be highly efficient. However, the part of PCA decomposition can be computationally demanding. Hence, only the first 15 sets of faces were used in each database.

Because of the long time it takes to calculate the PCA decomposition, this process was done in advance (offline), using the Signal and Image Processing Lab (SIPL) Matlab server. It should be clear that when the principal components for the training sets are known, applying the same process for the testing tests is easier.

It should also be noted that the k-NN algorithm is much less complex than SVM. It also shows better performance, so its results are presented here. Moreover, when solving the numerical optimization problem of SVM, convergence is not necessarily attained. In the attached Matlab file, the results for both methods can be obtained easily.

\subsection{AT\&T (ORL) database}

AT\&T (ORL) database \cite{ORL} contains 10 different images (92 x 112) each for 40 distinct subjects. Images in this database were taken at different times varying the lighting, facial expression and facial details (glasses/no glasses). All the images were taken against a dark homogeneous background with the subjects in an upright, frontal position (with tolerance for some side movement). 

\begin{figure}[h!]
\begin{center}
\includegraphics[scale=0.8]{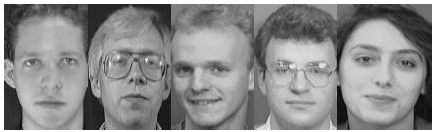}
\end{center}
\caption{Sample faces from AT\&T (ORL)}
\label{ORL}
\end{figure}

The pictures were taken between April 1992 and April 1994, and they are 8-bit grayscale in PGM format. Sample images of this dataset are shown in figure \ref{ORL}.

In Figure \ref{orl_results} below, the performance of the algorithm for the AT\&T (ORL) database is given. It can be seen that good results are obtained. It should be noted that some anomaly is seen (the recognition rate for training size of 6 should be higher than for training size of 5), but it is reasonable to assume that averaging on many sets would `smooth' this. This averaging could not be done due to limited computational resources.

\begin{figure}[h!]
\begin{center}
\includegraphics[scale=0.7]{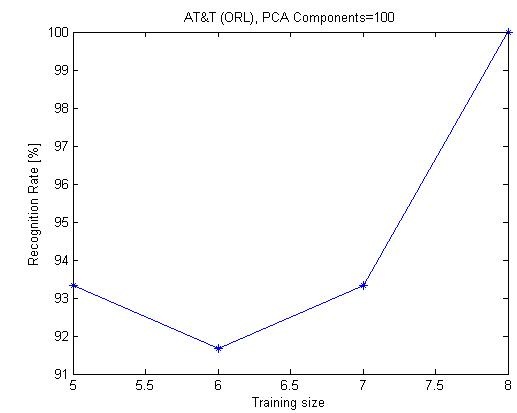}
\end{center}
\caption{Classification results for AT\&T (ORL)}
\label{orl_results}
\end{figure}

\subsection{Essex Grimace database}

\begin{figure}[h!]
\begin{center}
\includegraphics[scale=0.8]{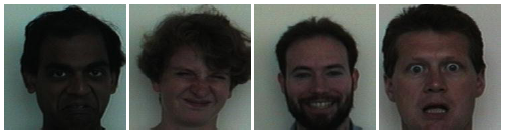}
\end{center}
\caption{Sample faces from Essex Grimace}
\label{essex}
\end{figure}

Essex Grimace database \cite{essex} contains a sequence of 20 images (180 x 200) each for 18 individuals consisting of male and female subjects, taken with a fixed camera. During the sequence, the subject moves his/her head and makes grimaces which get more extreme towards the end of the sequence. Images are taken against a plain background, with very little variation in illumination. Sample images of this database are shown in figure \ref{essex}.

In Figure \ref{grimace_results} below, the performance of the algorithm for the Grimace database is given.

\begin{figure}[h!]
\begin{center}
\includegraphics[scale=0.7]{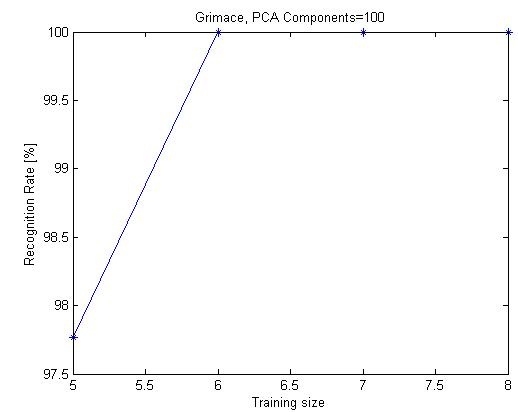}
\end{center}
\caption{Classification results for Grimace}
\label{grimace_results}
\end{figure}

It can be seen that even when less than half of the images are used as training set, the results are very good. This can be related to the fact that each picture in this database includes the whole face. Moreover, the background is homogeneous. The algorithm shows good results despite the fact that there are different face expressions.

\subsection{Georgia-Tech database}

\begin{figure}[h!]
\begin{center}
\includegraphics[scale=0.8]{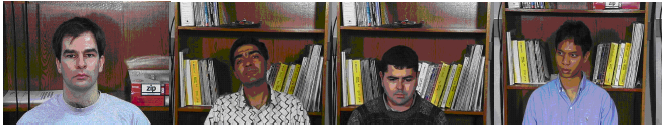}
\end{center}
\caption{Sample faces from Georgia-Tech}
\label{gt}
\end{figure}

Georgia Tech face database \cite{georgia} contains images of 50 taken during 1999. All people in the database are represented by 15 color JPEG images with cluttered background taken at resolution 640x480 pixels. The average size of the faces in these images is 150x150 pixels. The pictures show frontal and/or tilted faces with different facial expressions, lighting conditions and scale. Sample images can be seen in Figure \ref{gt}.

In Figure \ref{gt_results} below, the performance of the algorithm for the Georgia-Tech database is given.

\begin{figure}[h!]
\begin{center}
\includegraphics[scale=0.7]{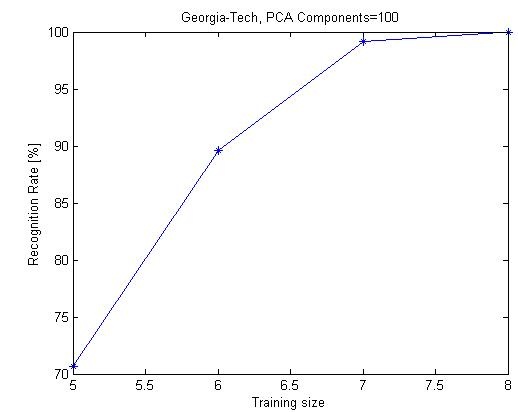}
\end{center}
\caption{Classification results for Georgia-Tech}
\label{gt_results}
\end{figure}

This database poses a challenge, since the pictures are not focused on faces. Moreover, the background is not homogeneous and the shadow of the head can been easily in each picture. Despite this issue, using even less than half of the pictures as training set provides great results.

\subsection{Comparison to known results}

\begin{figure}[h!]
\begin{center}
\includegraphics[scale=1]{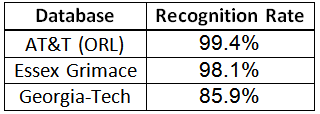}
\end{center}
\caption{Classification results from \cite{orig_paper}}
\label{orig_results}
\end{figure}

In \cite{orig_paper}, which is described in section \ref{orig_algorithm}, 6 images were used as training set for AT\&T (ORL), 12 for Grimace and 9 for Georgia-Tech database. The (averaged) results of this algorithm are given in Figure \ref{orig_results}.

The results are similar to our algorithm, whereas in the case of Georgia-Tech database our algorithm outperforms the algorithm given in \cite{orig_paper}. This can be explained by the challenging images of this database - the approach in \cite{orig_paper} produces quantized version of the same image to detect the face's curves, but it is not so efficient for this kind of pictures.

\begin{figure}[h!]
\begin{center}
\includegraphics[scale=0.6]{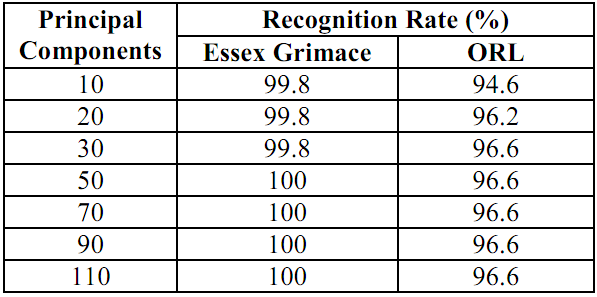}
\end{center}
\caption{Classification results from \cite{pca_paper}}
\label{pca_paper_results}
\end{figure}

In \cite{pca_paper}, which is described in section \ref{pca_algorithm}, 5 images per subject for AT\&T (ORL) and 8 images per subject for Grimace were used as training set (Georgia-Tech database was not used). The results are presented in Figure \ref{pca_paper_results}.

Our results are very similar, and in the case of Grimace they are even better.

In all the algorithms, the most difficult images to classify were those in which the subject's face is not directed at the camera. Indeed, it is reasonable to assume that the curves of the face are easier captured when the full face is directed to the camera.

\clearpage
\section{Matlab code}
\label{matlab_code}

This report is accompanied by a Matlab code which can produce the results presented before, along with many more. One needs just to define which database he wants to use, what is the training set size (per each subject) and how many components (of PCA) should be used.

As described above, the PCA decomposition process was done in advance, so its results for 50 to 120 components are saved under the relevant `mat' files. It was also done for each database and for each training set size (5 to 8), so the results are given very quickly. It also exempt us from the need to save all the databases.

SVM classification is disabled by default, since its performance are worse than k-NN. Moreover, it is computationally demanding and the optimization problem involved in this classification method does not necessarily converge, so the results of this method should be taken with caution.

\begin{figure}[h!]
\begin{center}
\includegraphics[scale=0.8]{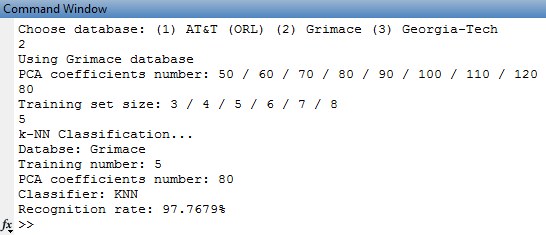}
\end{center}
\caption{Output results from Matlab - Example}
\label{matlab_example}
\end{figure}

You should just run "runme.m" file, after adding the main directory to Matlab path. More instructions are given in this file. Some output which was produced using this code is shown in Figure \ref{matlab_example}.

\clearpage
\section{Conclusions}
\label{conclusions}

In this project we demonstrated the ability of curevelet transform to assist in face recognition. It required learning about this transform and about common classification methods, such as k-NN and SVM. The main part of this project was the proper use of the curvelet transform coefficients as an input to a majority vote classifier, which was constructed according to the algorithm which was introduced in this project.

The initial idea was to implement one of the algorithms which were reviewed in \ref{known_algorithms}, but while looking for related material, the idea to build another algorithm which is based on these two appeared to be more challenging. Moreover, the main algorithm did not require the use of SVM classifier, and building such a classifier was also a significant part of the project, though k-NN classifier obtained better results.

Some implementation issues were considered. First, due to limited memory, we could use only the first 15 sets of faces in each dataset. Moreover, the PCA process takes a lot of time, so this process was done offline, and the PCA coefficients were saved. It should also be noted that SVM classifier deals usually with only two classes, so its adaptation to multi-class classification was also a significant part of the project. 

For improving the results of the algorithm, it is suggested to add some pre-processing steps. The first one should crop the picture so only the face would be shown (face detection). This could improve the results especially for Georgia-Tech database. For this task, it would be a good idea to use also the color information (instead of converting the images to grayscale) of the image. For example, we can determine parts of the face according to their colors (e.g., lips, cheeks). In addition, it would be advisable to examine another version of multi-class SVM, the version where each sample is compared against another one. However, this would add high computational complexity to the algorithm.

\clearpage
\appendix

\section{Appendix: k-NN}
\label{k-NN_section}

The K-nearest-neighbour algorithm \cite{wiki:KNN} is one of the simplest classification algorithms. The training examples are vectors in a multidimensional feature space, each with a class label. The training phase of the algorithm consists only of storing the feature vectors and class labels of the training samples.

In the classification phase, $k$ is a user-defined constant, and an unlabelled vector (a \emph{query} or \emph{test point}) is classified by assigning the label which is most frequent among the K training samples nearest to that query point. Many metrics can be used for measuring the distance between two points. Such metrics are the Euclidian (standard 2-norm) metric and the Gaussian metric $\left\| {\bf{x} - \bf{y}} \right\|=\exp \left( { - {{{{\left\| {{\bf{x}} - {\bf{y}}} \right\|}^2}} \over {2 \sigma^2}}} \right)$ with predefined values for $\sigma$ and K.

\begin{figure}[h!]
\begin{center}
\includegraphics[scale=0.7]{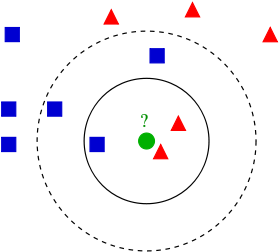}
\caption[Example of K-NN classification]{Example of K-NN classification. If K=1, the green circle will be classified to the same class as the blue triangles} \label{knn_classification}
\end{center}
\end{figure}

Obviously, this algorithm is very simple, where only the distances between any query point to all the points in the training set need to be calculated. It also offers pretty good results despite of its low complexity.

\section{Appendix: SVM}
\label{SVM_section}

\begin{figure}[h!]
\begin{center}
\includegraphics[scale=0.3]{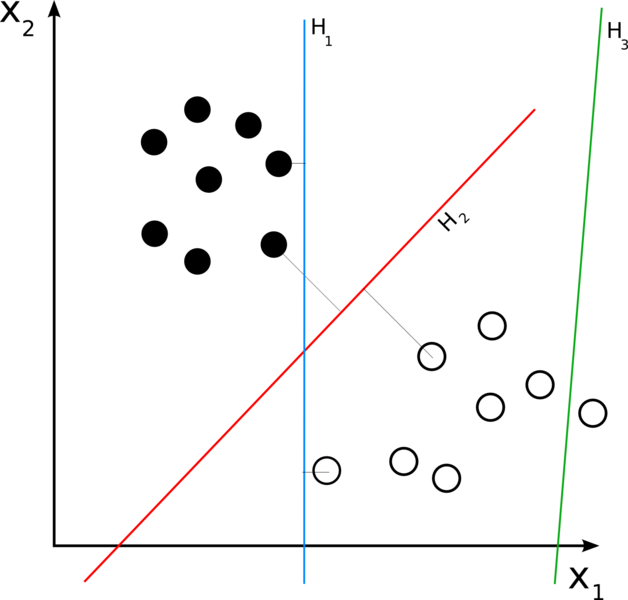}
\caption[Example of SVM classification]{H3 doesn't separate the 2 classes. H1 does, with a small margin and H2 with the maximum margin.} \label{svm_example}
\end{center}
\end{figure}

An SVM classifier \cite{SVM} tries to construct a separating hyper-plane between two groups of samples, in a way that the distance between the sets to the hyper-plane is maximal. It is demonstrated in Figure \ref{svm_example}.

As implied in the way that SVM classifier works, it separates only two groups, so in case there are more than two groups, some generalization is needed. One-against-all (OAA) SVMs were first introduced by Vladimir Vapnik in 1995. The initial formulation of the one-against-all method required unanimity among all SVMs: a data point would be classified under a certain class if and only if that class's SVM accepted it and all other classes' SVMs rejected it.

\begin{figure}[h!]
\begin{center}
\includegraphics[scale=3]{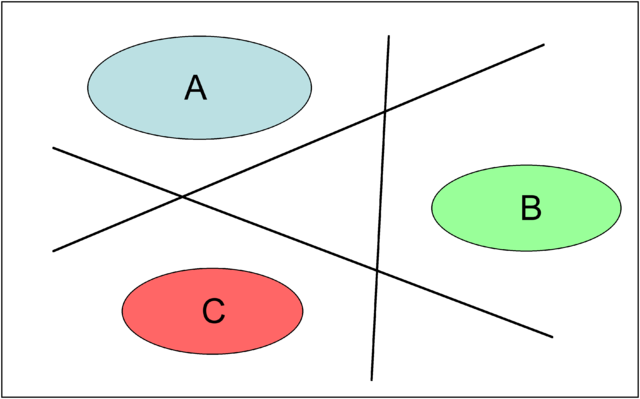}
\caption{Diagram of binary OAA region boundaries on a basic problem} 
\label{OAA_example}
\end{center}
\end{figure}

OAA method was used in this project. This method is depicted in Figure \ref{OAA_example}.

\clearpage
\listoffigures
\clearpage
\bibliography{project}

\begin{thebibliography}{10}

\bibitem{Maj07}
Angshul Majumdar and Arusharka Bhattacharya.
\newblock {A Comparative Study in Wavelets, Curvelets and Contourlets as
  Feature Sets for Pattern Recognition}.
\newblock In {\em 4th International Symposium on Visual Computing}, pages
  297--306, December 2008.

\bibitem{wiki:wt}
Wikipedia.
\newblock Wavelet transform --- {W}ikipedia{,} the free encyclopedia, 2004.
\newblock [Online; accessed 16-Feb-2011].

\bibitem{curvelets}
Emmanuel~J. Candµes and David~L. Donoho.
\newblock Curvelets - a surprisingly effective nonadaptive representation for
  objects with edges.
\newblock In {\em Curves and Surfaces, IV ed.}, 1999.

\bibitem{FDCT}
Emmanuel Candes, Laurent Demanet, David Donoho, and Lexing Ying.
\newblock {Fast Discrete Curvelet Transforms}.
\newblock In {\em Multiscale Modeling and Simulation}, pages 861--899, 2006.

\bibitem{orig_paper}
Tanaya Mandal, Angshul Majumdar, and Q.~Wu.
\newblock Face recognition by curvelet based feature extraction.
\newblock In {\em Image Analysis and Recognition}, volume 4633 of {\em Lecture
  Notes in Computer Science}, pages 806--817. Springer Berlin / Heidelberg,
  2007.

\bibitem{pca_paper}
T.~Mandal and Q.~Wu.
\newblock Face recognition using curvelet based {PCA}.
\newblock In {\em International Conference for Pattern Recognition - ICPR},
  pages 1--4, December 2008.

\bibitem{5583642}
Xianxing Wu and Jieyu Zhao.
\newblock Curvelet feature extraction for face recognition and facial
  expression recognition.
\newblock In {\em Natural Computation (ICNC), 2010 Sixth International
  Conference on}, volume~3, pages 1212 --1216, 2010.

\bibitem{Pinales}
Acosta-Reyes J.J. Salazar-Garibay A. Jaime-Rivas~R. Ruiz-Pinales, J.
\newblock Shift invariant support vector machines face recognition system.
\newblock In {\em Transactions on Engineering, Computing And Technology 16},
  pages 161--171, 2006.

\bibitem{ORL}
{Database by AT\&T Laboratories, \newline}
  {\href{http://www.cl.cam.ac.uk/research/dtg/attarchive/facedatabase.html}{http://www.cl.cam.ac.uk/research/dtg/attarchive/facedatabase.html}}.

\bibitem{essex}
{Database by University of Essex, \newline}
  {\href{http://cswww.essex.ac.uk/mv/allfaces/grimace.html}{http://cswww.essex.ac.uk/mv/allfaces/grimace.html}}.

\bibitem{georgia}
{Database by Georgia Institute of Technology, \newline}
  {\href{{http://www.anefian.com/face\_reco.htm}}{{http://www.anefian.com/face\_reco.htm}}}.

\bibitem{wiki:KNN}
{K-nearest neighbor algorithm} --- {W}ikipedia{,} the free encyclopedia, 2010.
\newblock [Online; accessed 16-Feb-2011].

\bibitem{SVM}
{Lecture notes in Machine Learning by Meir, Shimkin and Manor, Technion, IIT},
  2009.

\end{thebibliography}

\end{document}